\documentclass[letterpaper, 10 pt, conference]{ieeeconf}  

\usepackage[utf8]{inputenc}
\usepackage[T1]{fontenc}
\usepackage{verbatim}
\usepackage{graphicx}
\usepackage{caption}
\usepackage{subcaption}
\usepackage{wrapfig}
\usepackage{amsmath}
\usepackage[
    natbib=true,
    style=numeric,
    sorting=none
    ]{biblatex}
\usepackage[dvipsnames]{xcolor}

\newcommand{\customfigBracket}[1]{\textit{(Figure \ref{#1})}}

\newcommand\tm[1]{\textbf{\textcolor{blue}{(TM: #1)}}}

\IEEEoverridecommandlockouts                              

\begin{document}
\title{\LARGE \bf
	Eversion Robots for Mapping Radiation in Pipes
}

\author{Thomas Mack, Mohammed Al-Dubooni, Kaspar Althoefer, \textit{Senior Member, IEEE}
    \thanks{The first author is funded by an iCASE EPSRC PhD studentship.}
    \thanks{The second author is funded by an NDA PhD bursary}
    \thanks{Authors are with the Centre for Advanced Robotics @ Queen Mary, School of Engineering and Materials Science, Queen Mary University of London, United Kingdom.
        }%
    }

\maketitle
\thispagestyle{empty}
\pagestyle{empty}

\begin{abstract}
    A system and testing rig were designed and built to simulate the use of an eversion robot equipped with a radiation sensor to characterise an irradiated pipe prior to decommissioning. The magnets were used as dummy radiation sources which were detected by a hall effect sensor mounted in the interior of the robot. The robot successfully navigated a simple structure with sharp 45° and 90° swept bends as well as constrictions that were used to model partial blockages.
\end{abstract}

\section{Introduction}

    The substitution of humans with robots in hazardous environments is becoming commonplace, especially in the nuclear decommissioning industry where the constant presence of radioactive materials and waste warrants extreme safety precautions for almost all aspects of operation. 
    
    The exploration and characterisation of potentially irradiated spaces is one such area for which robotic solutions have been developed \parencite{mobileNuclearRobot}. However, the deployment of in-situ inspection solutions to the pipes and ducts that riddle these facilities is made difficult due to the small access ports. Their uses ranged from ventilation to waste drainage and many are positioned in hard to reach areas that make characterising them from the outside extremely difficult or impossible.
    
    Many robotic devices have already been developed for the inspection of pipes \parencite{pipeReview}, but most rigid solutions are highly restricted by their size, making them unsuitable for many of the thinner pipes. They also risk spreading contaminants through the pipe and must be decontaminated before reuse or disposed of leading to unwanted extra costs.

\begin{figure}[h]
            \centering
            \includegraphics[width=0.45\textwidth]{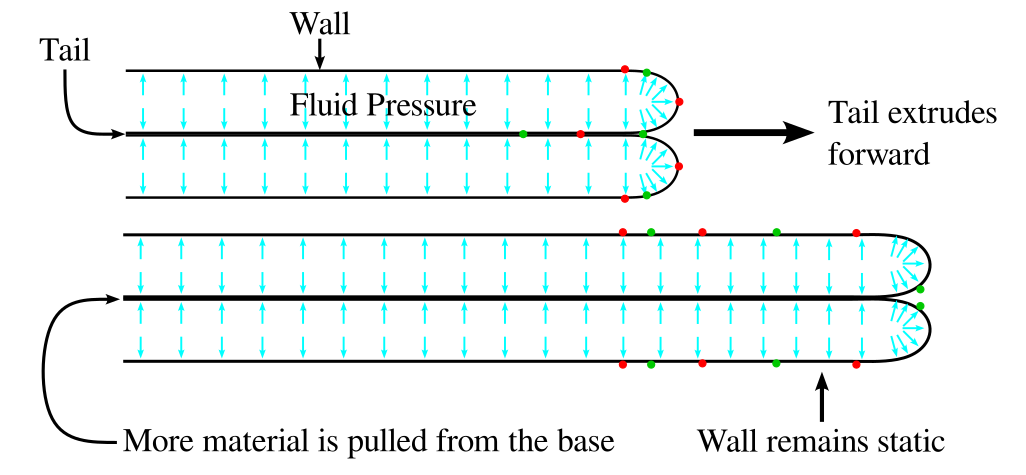}
            \caption{\small Basic construction and operation of the eversion robot
            }
            \label{fig:explanation}
        \end{figure}

    Eversion robots \textit{(Figure \ref{fig:explanation})} - a subset of soft robots - hold significant advantages for this kind of environment. They consist of an inverted sleeve, usually made from fabric or plastic. When inflated, the sleeve grows forward, pulling a tail of new material from the base. The outer walls remain static with the environment and do not exert frictional forces. As a result, they have medical applications in and outside the body where delicacy is required \parencite{colonoscope}.
    
    The pipe provides a guide for the robot to travel along, reducing the need for complex control and transforming the mapping into a one-dimensional problem, assuming an absence of forks. They also allow a robot that fills the pipe to retract without any extra assistance such as the retraction device in the cap created by \textcite{retractionDevice}. Due to the soft nature of the robot, it will also able to squeeze past some partial blockages. 

    Adding sensors to the tip of eversion robots is a presently studied challenge due to the material continuously moving as the robot extends. Most caps are made from rigid materials and fully encase the tip \parencite{retractionDevice, vineRobots}, limiting the size of the aperture they can fit through and undermining the eversion robot's ability to squeeze through spaces smaller than itself. A soft, fabric solution exists \parencite{softCap}, but that can have difficulties remaining in place while the robot retracts.

\section{Proposal}
   The main focus of characterising pipes in a nuclear environment is locating contaminated areas. The thin walls of an eversion robot would be easily penetrated by beta and gamma radiation. Thus, small radiation sensors could be placed within the robot on the end of the tail instead of mounting them on the tip. They would travel down the full length of the robot as it everts, measuring the distribution of radiation. This will remove the need for a cap and not limit the robot from passing through tight spaces. While this will prevent a camera from being mounted, it will allow the robot to continue to measure radiation levels past where it is feasible to take a camera on a rigid cap.

   To simulate this, we replaced the radiation sources with magnets that we detected with a hall-effect sensor. They were placed on a reconfigurable course of pipework for the robot to pass through and detect their locations. 
\section{Experimental Study}
    \subsection{Mock-up Environment}
        Simple courses of 55mm diameter plastic tubing were constructed to simulate pipes that could be found in a nuclear facility. They contained sharp 45° and swept 90° bends as well as some constrictions as small as 40mm. There were no splits in the path as the robot is, so far, unable to actively direct itself in a more complex network. Two sets of magnets were placed along the pipe for detection.

            \begin{figure}[h]
                \centering
                \includegraphics[width=0.4\textwidth]{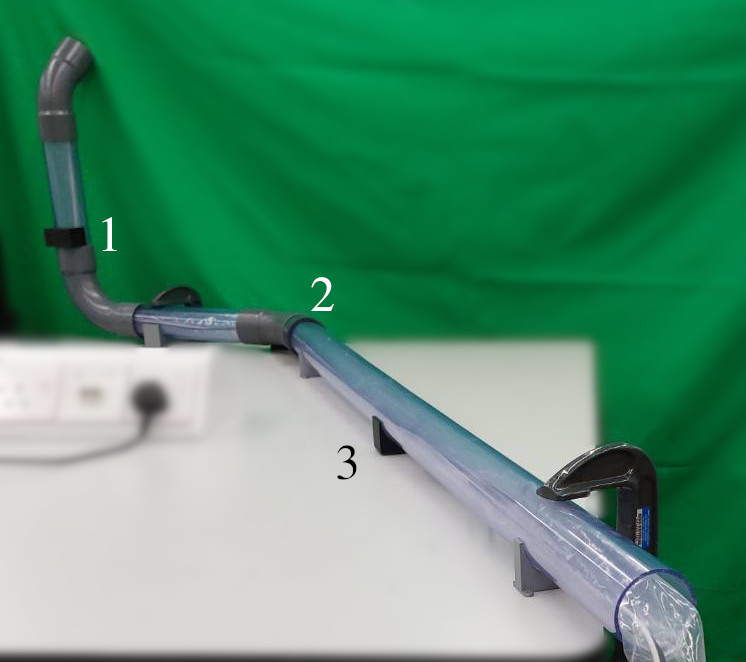}
                \caption{\small The 55mm pipes could be taken apart and placed in different configurations for the robot to traverse and map the position of the magnets imitating radiation sources. 1, 3: The two magnets placed on the side of the pipe encased in black plastic. 2: The constriction.}
                \label{fig:environment}
            \end{figure}
            
    \subsection{Eversion Robot Construction}
        The first robot was constructed from Rip-Stop Nylon fabric and the second robot from Polythene lay-flat tubing.
        
        The fabric tubing was sewn from two strips into a 5m long, 60mm diameter tube. Excess fabric on the seams was cut as short as possible and then heat sealed with vinyl to eliminate air leaks. One of the open ends was sewn and similarly sealed.

        The plastic tubing could just be heat sealed at one end to form the shape we needed. However, it was not available in a size that fully filled our test rig, so we used the next smallest version.

        A tendon was tied to the sealed end of both sleeves, enabling controlled extension and retraction. It was attached to a 3D-printed, hand-cranked drum which can be used to infer the robot's extension using a shaft encoder. A hall effect sensor was also attached at the sealed end with a wire leading out of the base of the robot. A clamp was placed around the end of the eversion robot to minimise air leaks from the seal around the tendon and wire.

        A small ROS package was written to read the shaft encoder and the hall effect sensor to plot them in real time.

\section{Results}
    \begin{figure}[h]
        \centering
        \includegraphics[width=0.4\textwidth]{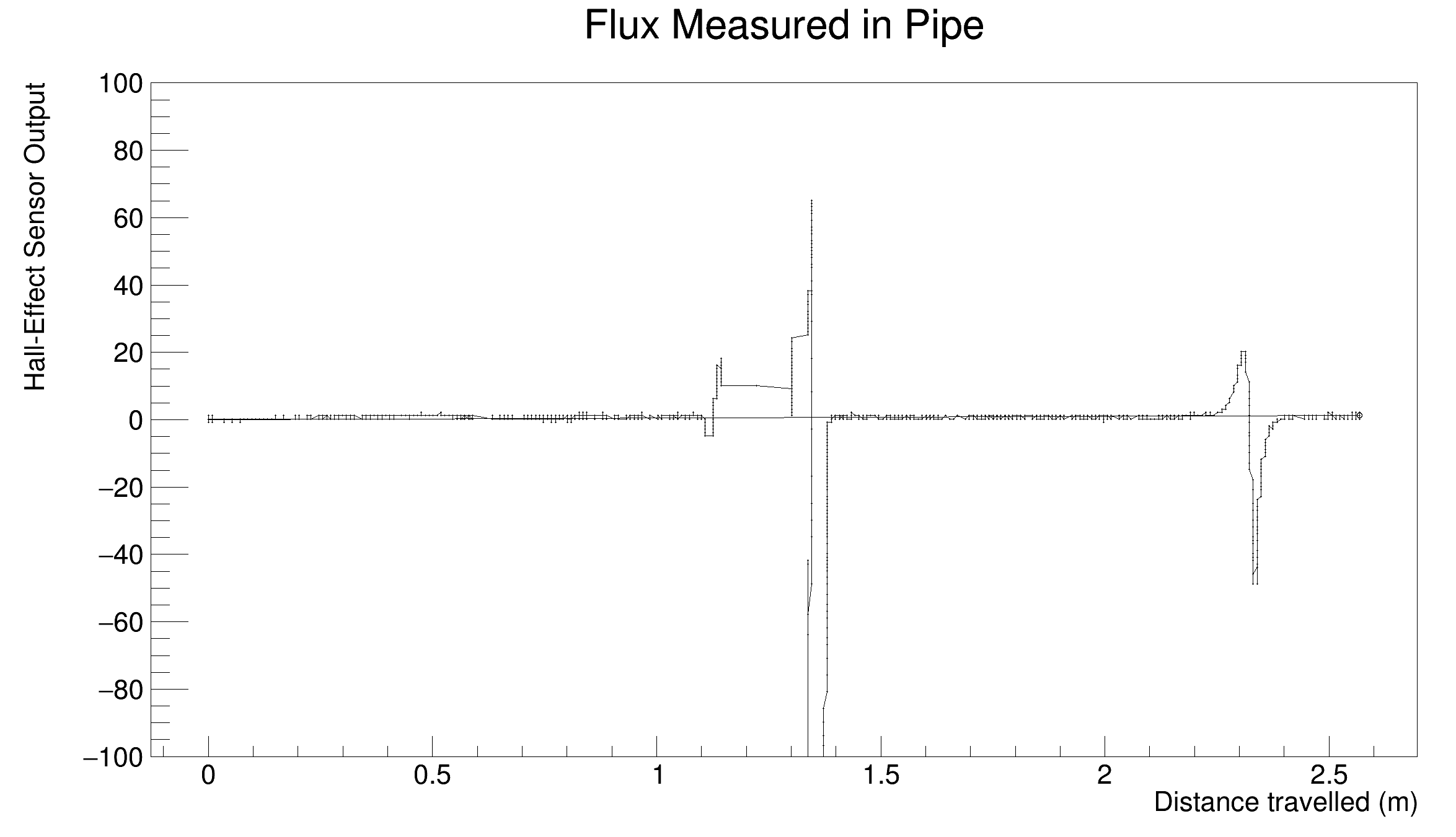}
        \caption{\small A graph produced by the plastic robot in the experimental setup shown in \textit{Figure \ref{fig:environment}}}
        \label{fig:graphs}
    \end{figure}
    
    Both robots were able to evert through the straight piece of pipe and the constrictions without any problems, but the fabric robot was unable to extend past the 45° bends. It was noted that due to manufacturing imperfections, there were multiple small leaks on the seams causing pressure loss, and the vinyl sealing created a stiff section which may have hindered extension.

    The plastic robot was uniform and almost fully airtight, save for the clamped base where the wire and tendon had to pass through. It successfully transported the hall-effect sensor through every course of pipes that we assembled. However, it was not possible to cleanly retract it as it did not fill the pipe, which caused it to buckle and bend considerably.

    Clear spikes \textit{(Figure \ref{fig:graphs})} in the graphs were produced by the hall effect sensor which could be used to measure where on the pipe the magnets were placed. However, there was always a considerable amount of distance before the sensor encountered a magnet because it had to travel half the full length of the robot before it entered the pipe.
    
\section{Conclusion}
    The system has been shown to work as a proof of concept for mapping the radiation levels in simple pipe structures with no branching paths. The plastic lay-flat tubing worked best for everting through constrictions and sharp bends as it was able to retain more pressure than the fabric self-made robots.

    In future, the entire assembly will be encased in an airtight chamber to reduce friction on the tendon and cable as well as let the entire robot be retracted and rolled up on the drum. A stepper motor will be used to rotate the drum instead so the mapping can be automated. The plastic lay-flat tubing will also be tested with pipe of the correct diameter.
\addtolength{\textheight}{-12cm}

\AtNextBibliography{\small}
\printbibliography

\end{document}